\documentclass[preprints,article,accept,pdftex,moreauthors]{Definitions/mdpi} 

\firstpage{1} 
\pubvolume{1}
\issuenum{1}
\articlenumber{0}
\pubyear{2023}
\copyrightyear{2023}
\datereceived{6 November 2023} 
\daterevised{22 November 2023} 
\dateaccepted{23 November 2023} 
\datepublished{ } 
\hreflink{} 

\usepackage{palatino}

\usepackage{graphicx}
\usepackage{caption}
\usepackage{subcaption}
\usepackage{xcolor}
\usepackage{colortbl}

\makeatletter
\def\UrlAlphabet{%
\do\a\do\b\do\c\do\d\do\e\do\f\do\g\do\h\do\i\do\j%
\do\k\do\l\do\m\do\n\do\o\do\p\do\q\do\r\do\s\do\t%
\do\u\do\v\do\w\do\x\do\y\do\z\do\A\do\B\do\C\do\D%
\do\E\do\F\do\G\do\H\do\I\do\J\do\K\do\L\do\M\do\N%
\do\O\do\P\do\Q\do\R\do\S\do\T\do\U\do\V\do\W\do\X%
\do\Y\do\Z}
\def\UrlDigits{\do\1\do\2\do\3\do\4\do\5\do\6\do\7\do\8\do\9\do\0}
\g@addto@macro{\UrlBreaks}{\UrlOrds}
\g@addto@macro{\UrlBreaks}{\UrlAlphabet}
\g@addto@macro{\UrlBreaks}{\UrlDigits}
\makeatother

\Title{adaptMLLM: Fine-Tuning Multilingual Language Models on Low-Resource Languages with Integrated LLM Playgrounds}

\TitleCitation{adaptMLLM: Fine-Tuning Multilingual Language Models on Low-Resource Languages with Integrated LLM Playgrounds}

\Author{Séamus Lankford 
 $^{1,2,}$*$^{,\dagger,\ddagger}$\orcidA{}, Haithem Afli $^{2,\ddagger}$ and Andy Way $^{1,\ddagger}$}

\AuthorNames{Séamus Lankford, Haithem Afli and Andy Way}

\AuthorCitation{Lankford, S.; Afli, H.; Way, A.}

\address{%
$^{1}$ \quad ADAPT Centre, School of Computing, Dublin City University, D09 DXA0 Dublin, Ireland
\\
$^{2}$ \quad Department of Computer Science, Munster Technological University, T12 P928 Cork, Ireland
}

\corres{Correspondence: seamus.lankford@mtu.ie 
}

\firstnote{Current address: ADAPT Centre, School of Computing, Dublin City University, D09 DXA0 Dublin, Ireland} 
\secondnote{These authors contributed equally to this work. }

\abstract{The advent of Multilingual Language Models (MLLMs) and Large Language Models \linebreak  (LLMs) has spawned innovation in many areas of natural language processing. Despite the exciting potential of this technology, its impact on developing high-quality Machine Translation (MT) outputs for low-resource languages remains relatively under-explored. Furthermore, an open-source application, dedicated to both fine-tuning MLLMs and managing the complete MT workflow for low-resources languages, remains unavailable. We aim to address these imbalances through the development of adaptMLLM, which streamlines all processes involved in the fine-tuning of MLLMs for MT. This open-source application is tailored for developers, translators, and users  who are engaged in MT. It is particularly useful for newcomers to the field, as it significantly streamlines the configuration of the development environment. An intuitive interface allows for easy customisation of hyperparameters, and the application offers a range of metrics for model evaluation and the capability to deploy models as a translation service directly within the application. As a multilingual tool, we used adaptMLLM to fine-tune models for two low-resource language pairs: English to Irish (EN$~\leftrightarrow~$GA) and English to Marathi (EN$~\leftrightarrow~$MR). Compared with baselines from the LoResMT2021 Shared Task, the adaptMLLM system demonstrated significant improvements. In the EN$~\rightarrow~$GA direction, an improvement of 5.2 BLEU points was observed and an increase of 40.5 BLEU points was recorded in the GA$~\rightarrow~$EN direction representing relative improvements of 14\% and 117\%, respectively. Significant improvements in the translation performance of the EN$~\leftrightarrow~$MR pair were also observed notably in the MR$~\rightarrow~$EN direction with an increase of 21.3 BLEU points which corresponds to a relative improvement of 68\%. Finally, a fine-grained human evaluation of the MLLM output on the EN$~\rightarrow~$GA pair was conducted using the Multidimensional Quality Metrics and Scalar Quality Metrics error taxonomies. The application and models are freely available. 
}


\begin{document}


\section{Introduction}

\label{Introduction}

Large Language Models (LLMs), are AI models that use deep learning techniques to generate human-like text. These models are trained on vast amounts of text data, often using unsupervised learning, to learn the patterns and relationships within language. This results in models that can generate text which is often indistinguishable from text written by a human.

The excitement surrounding LLMs stems from their potential to revolutionise many fields, from language translation~\citep{costa2022no} and content generation~\citep{brown2020language} to chatbots e.g. \url{https://openai.com/blog/chatgpt} (accessed on 28 November 2023) and virtual assistants e.g. \url{https://genie.stanford.edu/} (accessed on 28 November 2023). With their ability to understand natural language and generate complex responses, LLMs have the potential to enhance human communication and productivity in ways that were previously unimaginable. LLMs can also be used in creative applications, such as generating music e.g. \url{https://soundraw.io/} (accessed on 28 November 2023) or art e.g. \url{https://labs.openai.com/} (accessed on 28 November 2023).

No Language Left Behind (NLLB)~\citep{costa2022no} represents a groundbreaking AI project in the area of Multilingual Language Models (MLLMs). The project has released open-source models proficient in delivering high-quality translations across 200 languages and has enhanced translations for low-resource languages on platforms like Facebook and Instagram. The NLLB-200 model, integrated into the Wikimedia Foundation's Content Translation tool, aids Wikipedia editors in translating content into their preferred languages. These editors can now more effectively translate articles from lesser-known languages, such as Luganda and Icelandic, enriching Wikipedia's language diversity. The open-sourced nature of the NLLB-200 model also empowers the research community and Wikipedia editor groups to expand upon their findings.

When building LLMs, the focus is on designing and training the model architecture. This involves selecting the appropriate neural network architecture and hyperparameters, as well as deciding on the training data and optimisation techniques to use.

Tuning an MLLM or LLM, on the other hand, involves adjusting the parameters of the model to improve its performance on a specific task. In neural networks such as MLLMs and LLMs, the weights and biases are parameters that the network adjusts through training to minimise a cost function. This is performed by training the model on a task-specific dataset and adjusting the model's hyperparameters to optimise its performance. Tuning an MLLM can be a challenging task, as the model is often very complex and the training process can take a long time. Our paper concentrates on fine-tuning pre-built MLLMs to enhance machine translation (MT) with a particular focus on low-resource language pairs. 

The process of fine-tuning an MLLM involves several distinct stages which are broken down into individual steps. These steps include setting up the environment, preparing the dataset, parameterising and fine-tuning the chosen MLLM, and evaluating and deploying the model. This modular approach has proven to be effective in fine-tuning MLLMs, and we have structured our adaptMLLM application to cater for both developers and translators. In light of the environmental impact of developing and running large AI models~\citep{strubell-etal-2019-energy,henderson2020towards}, we also calculate carbon emissions in a ``green report''. It is envisaged that such a report will incentivise more responsible and sustainable model development.

A significant aspect of our research involves creating applications and models to address language technology challenges. Similar to our previous work, which focused on developing NMT models~\citep{lankford2023adaptNMT}, we hope this paper will be particularly helpful for those new to MT wishing to learn more about fine-tuning MLLMs.

Unlike many translation toolkits, our application does not use a command line interface. Instead, we have designed and fully implemented the interface in Google Colab, (\url{https://colab.research.google.com}, accessed on 28 November 2023) a cloud-hosted solution (\url{https://cloud.google.com}, accessed on 28 November 2023) that is more intuitive for both educational and research settings. Furthermore, our application provides Graphical User Interface (GUI) controls within adaptMLLM, enabling users to customise all key hyperparameters required for MLLMs.

Our application is designed to operate as a platform as a service (PaaS) cloud computing application, allowing for quick and efficient scaling of the infrastructure. Additionally, the deploy function allows for immediate deployment of trained models.

This paper is organised by initially presenting related work and background information on MLLMs and LLMs in Section \ref{related}. This is followed by a description of our datasets in Section \ref{approach}. The key features of the adaptMLLM architecture are discussed in Section \ref{aLLM} and an empirical evaluation of our trained models, including a human evaluation is carried out in Section \ref{sec:exp}. The system is discussed in Section \ref{sec:discussion} before drawing conclusions and describing future work in Section \ref{concl}.

\section{Related Work}
\label{related}

\subsection{Transformer Architecture}

After the attention mechanism was introduced, researchers naturally began to explore whether attention alone could handle the bulk of the translation task. Accordingly, Vaswani et al. proposed that ``attention is all you need'' in their Transformer architecture~\citep{10.5555/3295222.3295349}, which has achieved state-of-the-art (SOTA) performance on many natural language processing (NLP) benchmarks by exclusively using an attention mechanism, eliminating the need for recurrence and convolution, and enabling the employment of far simpler feed-forward neural networks.

In the context of our research, we have previously demonstrated that Transformer-based models deliver high-functioning models for the low-resource EN$~\rightarrow~$GA language pair~\citep{lankford-etal-2021-transformers}.

The default Transformer architecture follows an encoder--decoder structure generating its output without relying on recurrence and convolutions. The encoder's role is to convert an input sequence into a series of continuous representations, which are subsequently fed into a decoder. The decoder produces an output sequence by using the encoder's output in combination with the output generated by the decoder at the preceding time step.

\subsection{Multilingual Language Models---NLLB}

MT has become a significant area of research in AI with the aim of eliminating language barriers worldwide. However, the current focus is limited to a small number of languages, neglecting the vast majority of low-resource languages. In an effort to address this issue, the No Language Left Behind (NLLB) initiative was launched. This project aims to overcome the challenges of using MT for low-resource language translation by developing datasets and models that bridge the performance gap between low- and high-resource languages. The NLLB team has also created architectural and training enhancements tailored to support MT for low-resource languages. Their work is open source, (\url{https://github.com/facebookresearch/fairseq/tree/nllb}, accessed on 28 November 2023), and many of their models serve as baselines for fine-tuning with adaptMLLM.

\subsection{Large Language Models}\label{llm_background}

The increasing availability of large datasets provides the raw material for LLM training~\citep{radford2019language,conneau-etal-2020-unsupervised,winata-etal-2021-language},
enabling performance improvement on NLP tasks, which can learn from a wide variety of sources. 

Another key factor in driving the ubiquity of LLMs has been the growth in computational power dedicated to the domain. As a consequence, more powerful computers now have the capability to train LLMs on massive datasets which, in turn, has led to SOTA results on many common NLP tasks~\citep{devlin-etal-2019-bert}. New training algorithms developed through advancement in AI research has further boosted LLM performance~\citep{lepikhin2020gshard}. 

LLMs have the potential to improve the use of technology across a wide range of domains, among which include medicine, education and computational linguistics. In education, LLMs may be used for personalised student learning experiences~\citep{KASNECI2023102274}, while in the medical domain, analysing large amounts of medical files can assist doctors in treating patients~\citep{iftikhar2023docgpt}. 
Of particular interest to our research is the manner in which LLMs can be used within the realm of computational linguistics, more specifically in the field of MT. 

\subsubsection{GPT-J}

Transformers are increasingly the architecture of choice for NLP problems, replacing Recurrent Neural Networks (RNNs) such as Long Short-Term Memory (LSTM)~\citep{hochreiter1997long}. 

GPT-J is an open-source implementation of a particular class of LLMs known as Generative Pre-trained Transformer (GPT) models~\citep{radford2018improving}. GPT-J is a Transformer model trained using Wang's Mesh Transformer JAX (\url{https://github.com/kingoflolz/mesh-transformer-jax}, accessed on 28 November 2023). GPT-J-6B (\url{https://6b.eleuther.ai}, accessed on 28 November 2023) is an autoregressive language model, created by EleutherAI (\url{https://www.eleuther.ai}, accessed on 28 November 2023), with 6 billion trainable parameters. As an advanced alternative to OpenAI's GPT-3, it performs very well on a wide array of NLP tasks such as chat, summarisation, and question answering. 

\subsubsection{GPT-4}

The primary distinction between GPT-3.5 and GPT-4 (\url{https://openai.com/product/gpt-4}, accessed on 28 November 2023) is that while the former is a text-to-text model, the latter is more of a data-to-text model, exhibiting the ability to perform tasks that its predecessor could not. For example, GPT-4 is capable of processing visual input as part of a prompt, such as images or web pages, and can even generate text that explains the humour in memes. Consequently, GPT-4 can be classified as a ``multimodal model''. Furthermore, GPT-4 has a longer memory than its previous versions, with a short-term memory closer to 64,000 words, enabling it to maintain coherence during extended interactions. GPT-4 also enables users to select different personalities for the model's responses.

The number of parameters utilised in the training of GPT-4 has not been disclosed by OpenAI; however, other sources, such as AX Semantics (\url{https://en.ax-semantics.com/}, accessed on 28 November 2023), have estimated the number to be around 100 trillion. AX Semantics maintains that such a number makes the language model more akin to the functioning of the human brain with respect to language and logic (\url{https://https://en.ax-semantics.com/blog/gpt-4-and-whats-different-from-gpt-3/}, accessed on 28 November 2023). 
  
Additionally, GPT-4 outperformed GPT-3.5 in various standardised tests, such as the LSAT, SAT, Uniform Bar Exam, and GRE, and was shown to be 82\% less likely to respond when prompted inappropriately and 60\% less likely to generate false information~\citep{OpenAI2023GPT4TR}.
 
\subsubsection{BARD}

BARD (\url{https://bard.google.com/}, accessed on 28 November 2023) utilises a lightweight version of the Language Model for Dialogue Applications (LaMDA)~\citep{thoppilan2022lamda}, which is an AI engine developed by Google. BARD has two primary objectives: to ensure the accuracy of its responses and to integrate the benefits of AI into Google's everyday products. Google has a rich history of employing AI to improve the search experience for billions of users. Its earlier Transformer model, BERT (\url{https://github.com/google-research/bert}, accessed on 28 November 2023), was a breakthrough in comprehending the intricacies of human language. The company has since introduced MUM (\url{https://blog.google/products/search/introducing-mum/}, accessed on 28 November 2023), which is a thousand times more potent than BERT. Recent AI technologies like LaMDA, PaLM, Imagen, and MusicLM are building on these developments, creating new ways to interact with information from language and images to video and audio. Furthermore, in 2018, Google was one of the pioneering companies to release a set of AI principles (\url{https://ai.google/principles/}, accessed on 28 November 2023).

Apart from its own products, Google aims to assist developers in innovating with AI by simplifying and scaling the benefits of these advances. In the future, the company intends to create a suite of tools and APIs that will make it easier to build innovative applications with BARD and more generally with its AI.

\subsection{DeepSpeed}\label{deepspeed}

The advent of DeepSpeed~\citep{10.1145/3394486.3406703}, a free software library from Microsoft, was a significant breakthrough for researchers looking to implement and fine-tune MLLMs and LLMs with limited resources. Large model training, in terms of scale, speed, and cost, is now achievable for most people. Additionally, DeepSpeed's most recent Transformer kernel improvements enabled the DeepSpeed team to achieve SOTA performance, setting a new record for the fastest BERT~\citep{devlin-etal-2019-bert} pre-training.

For small teams, DeepSpeed's Zero Redundancy Optimizer (ZeRO) is particularly advantageous, providing fresh memory optimisation for large-scale distributed deep learning. With minor changes to a PyTorch model, DeepSpeed can improve the speed and scale of model training.

\subsection{HuggingFace}\label{hf}

The Hugging Face Transformers library (\url{https://github.com/huggingface/transformers}, accessed on 28 November 2023)~\citep{wolf-etal-2020-transformers} is an open-source software library that provides a wide range of pre-trained SOTA NLP models, including models for language modelling, question answering, text classification, and MT, among others.
 
The library is built on top of popular deep learning frameworks such as PyTorch (\url{https://github.com/pytorch/pytorch}, accessed on 28 November 2023) and TensorFlow, (\url{https://github.com/tensorflow/tensorflow}, accessed on 28 November 2023) and it provides a simple and consistent API for accessing pre-trained models and fine-tuning them for downstream tasks. The library also includes a set of tools for data preprocessing, model evaluation, and visualisation, which make it easier for researchers and developers to experiment with different NLP models and tasks.

The Hugging Face Transformers library has become one of the most popular and widely used NLP libraries in the industry and the research community, and it has been adopted by many companies and organisations to build NLP applications and systems.

\subsection{Human Evaluation}

Within the fields of NLP and MT, human evaluation is increasingly recognised as critical, often meriting its own specialised research track or workshop at leading conferences~\citep{humeval-2021-human}. This emphasis has spurred a wealth of studies focusing on human evaluation related to MT, proving especially valuable in assessing low-resource languages~\citep{bayon-sanchez-gijon-2019-evaluating,imankulova2019exploiting}.

A set of best practices for human evaluation in MT has emerged, detailed in a collection of suggested guidelines~\citep{laubli2020set}. Our study incorporates these guidelines, aligning with comparable EN~${\leftrightarrow}$~GA studies at the ADAPT centre. To enhance these guidelines, a detailed human analysis was conducted, employing both the Scalar Quality Metric (SQM)~\citep{freitag2021experts} and the Multidimensional Quality Metric (MQM)~\citep{lommel2014multidimensional} for a nuanced assessment. SQM and MQM, are both widely used in industry and academia, to evaluate the quality of machine-generated text.

SQM is a simple, single-number metric that is used to measure the overall MT quality. It is often used when a quick evaluation of the quality of the text is required.

MQM, on the other hand, is a more complex metric that measures the quality of the text across multiple dimensions such as fluency, adequacy, and coherence, to name a few. It provides a more comprehensive evaluation of MT by measuring the quality of the text across different aspects.

\section{Datasets}\label{approach}

\subsection{Language Pairs}
To evaluate the translation performance of adaptMLLM in fine-tuning MLLMs for low-resource languages, we had to choose suitable language pairs. Furthermore, appropriate datasets upon which we could benchmark our performance also had to be sourced. The EN~${\leftrightarrow}$~GA and EN~${\leftrightarrow}$~MR language pairs were selected since they fulfilled the criteria of low-resource languages.

The Irish language, also known as Irish Gaelic, is the first official language of the Republic of Ireland, and is also recognised as a minority language in Northern Ireland. According to the 2022 Irish census (\url{https://www.cso.ie/en/releasesandpublications/ep/p-cpsr/censusofpopulation2022-summaryresults/educationandirishlanguage/}, accessed on 28 November 2023), 1.87 million people in the Republic of Ireland reported being able to speak Irish to some degree, which represents 40.4\% of the population. Irish is also spoken by a small number of people in other countries, particularly in the United States, Canada, and Australia, as well as in Irish-speaking communities in other parts of the world. It is also one of the official languages of the European Union and a recognised minority language in Northern Ireland with an ISO code of ``GA'' (\url{https://www.iso.org/}, accessed on 28 November 2023).

The dominant language spoken in India's Maharashtra state is Marathi, with an ISO code of ``MR''. It has over 83 million speakers, and it is a member of the Indo-Aryan language family. Despite being spoken by a significant number of people, Marathi is considered to be relatively under-resourced when compared to other languages used \mbox{in the region.}

\subsection{Shared Task Datasets}

To benchmark the performance of our EN~${\leftrightarrow}$~GA models, trained using adaptMLLM, datasets from the LoResMT2021 Shared Task (\url{https://github.com/loresmt/loresmt-2021}, accessed on 28 November 2023)~\citep{ojha2021findings} were used. These datasets enabled the evaluation of adaptMLLM models, since the shared task focused on low-resource languages which included both the EN~${\leftrightarrow}$~GA pair and the EN~${\leftrightarrow}$~MR pair. Furthermore, using official datasets from a shared task enables our models' performance to be directly compared with models entered by other teams. 

Both datasets focused on the specific domain of translation of COVID-related data. A parallel corpus of EN~${\leftrightarrow}$~GA sentences concentrating on the COVID domain were mainly drawn from the Government of Ireland (\url{https://www.gov.ie/}, accessed on 28 November 2023) and the Health Service Executive (\url{https://www.hse.ie/}, accessed on 28 November 2023) websites. EN~${\leftrightarrow}$~MR parallel Covid sentences were extracted from the Government of India (\url{https://www.mygov.in/}, accessed on 28 November 2023) website, BBC Marathi (\url{https://www.bbc.com/marathi}, accessed on 28 November 2023) and online newspapers. A detailed breakdown of all sources is available in \cite{ojha2021findings}.

The datasets from the shared task provided 502 Irish and 500 Marathi validation sentences whereas 250 (GA~${\rightarrow}$~EN), 500 (EN~${\rightarrow}$~GA), and 500 (EN~${\leftrightarrow}$~MR) sentences were made available in the test datasets, i.e., exactly the same as our other experiments to allow direct comparison with previous work. Training data consisted of 20,933 lines of parallel data for the EN~${\leftrightarrow}$~MR language pair and 13,171 lines of parallel data were used to train the EN~${\leftrightarrow}$~GA models.

\section{Approach}\label{aLLM}
\label{disc}

\begin{figure}[H]
\begin{adjustwidth}{-\extralength}{0cm}
\centering 
 \includegraphics[width=17.5 cm]{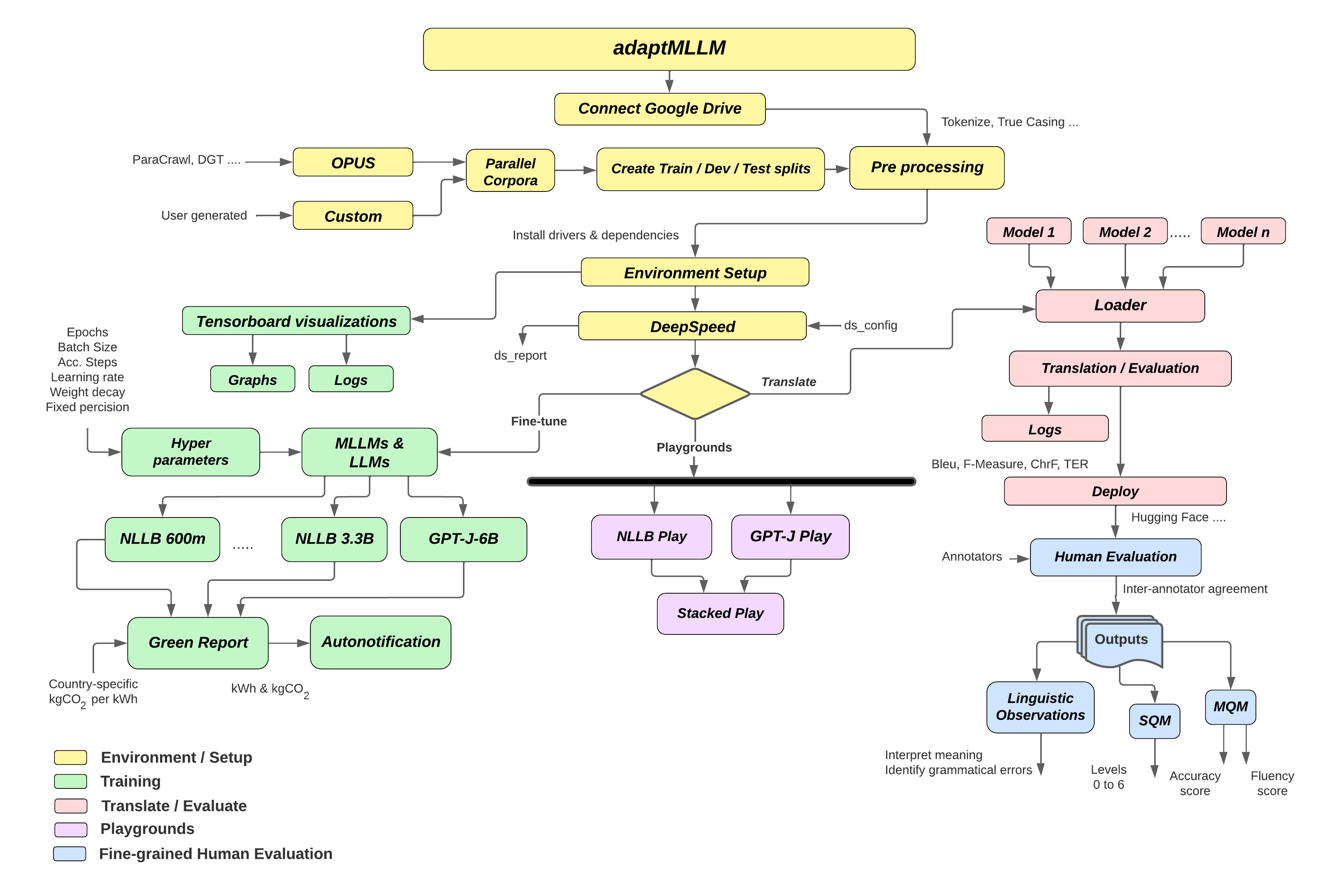}
\end{adjustwidth}
    \caption{Proposed architecture for adaptMLLM: a system for fine-tuning MLLMs.}
    \label{fig:arch}
\end{figure}

After outlining the background that gave rise to the creation of MLLMs and LLMs, we now introduce the adaptMLLM tool. This tool allows users to customise these components to their liking. Figure \ref{fig:arch} offers a high-level overview of the platform's system architecture.

The application is designed as an IPython notebook and employs Pytorch for model training. The utilisation of a Jupyter notebook format facilitates easy sharing within the AI community. Additionally, the challenge of configuring the proper development environment is substantially reduced, as all necessary packages are automatically downloaded while the application is running.

There are options to run the system for fine-tuning MLLMs, evaluating MLLM translation performance, testing LLM playgrounds and conducting a human evaluation of the translation performance. The application is run as a Colab instance on the Google Cloud.  Translation models are developed using aligned text corpora from both the original and the target languages. Tensorboard offers a live graphical representation of the training process of the model. The system is primarily employed for training models and functioning as a translation service, either of which can be chosen at run-time.

\begin{figure} [H]
\includegraphics[height=20.5cm]{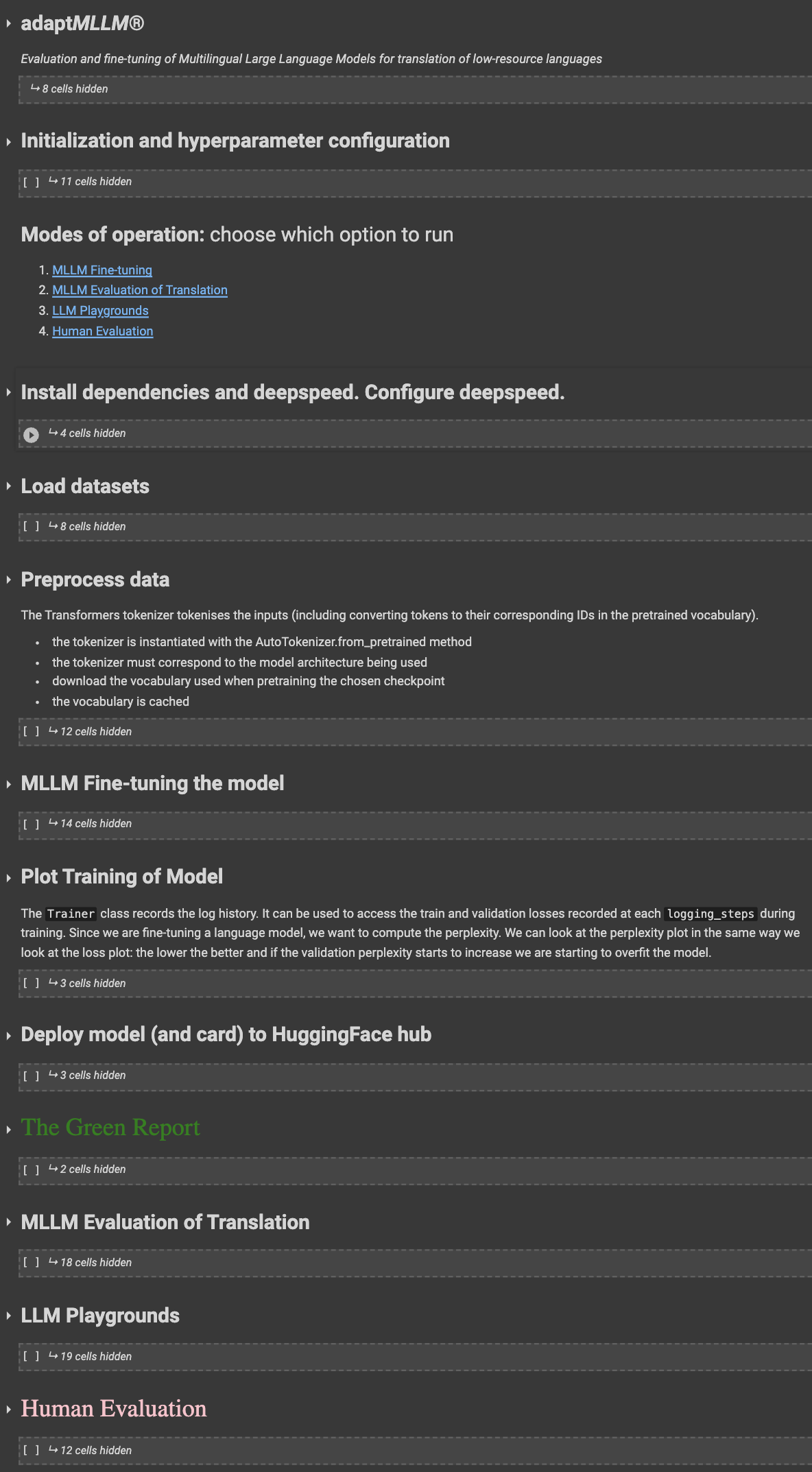}
  \caption[Overview of adaptMLLM]{Overview of adaptMLLM. Key areas include initialisation, menu of operation modes, loading and pre-processing, MLLM fine-tuning, visualisation, deployment, a green report, MLLM translation and evaluation, LLM playgrounds and human evaluation (cf. Section \ref{aLLM}).}
  \label{fig:features}
\end{figure}

The application is primarily run as a Google Colab application but may also be run as an Jupyter notebook. Given the ease of integrating Google drive storage into Colab, we have used adaptMLLM exclusively as a Google Colab application for our own experiments, some of which are described in Section \ref{sec:exp}. Key features of the notebook are highlighted in Figure \ref{fig:features}.

\subsection{Initialisation and Pre-Processing}

Initialisation enables connection to Google Drive to run experiments, automatic installation of Python, SentencePiece (\url{ https://github.com/google/sentencepiece}, accessed on 28 November 2023)~\citep{kudo2018sentencepiece}, Pytorch, HuggingFace Transformer's library (cf. Section \ref{hf}), and other libraries. 

The train, validation, and test splits for both source and target languages may be uploaded by the users. In cases where a user has not already created the required splits for model training, single source and target files may be uploaded. The necessary splits to form the training, validation, and test files will be automatically created based on the split ratio specified by the user. 

\vspace{-6pt}

\subsection{Modes of Operation}
There are several modes of operation, namely MLLM fine-tuning, evaluation of MLLM translation performance, experimentation with LLM playgrounds, and a human evaluation of the translation output. 

With MLLM fine-tuning, the application develops models using Google's GPU-based cloud platform. For a monthly subscription, the Google Colab Pro+ is a prerequisite since fine-tuning demands access to high-end GPU and compute resources.  

Apart from low-cost access to a high-spec infrastructure, model development on the Google Cloud is also recommended given the platform uses 100\% renewables~\citep{lacoste2019quantifying}. This has emerged as an economical choice for practitioners in the field of low-resource languages, as the creation of smaller models involves reduced training times.

\subsection{Fine-Tuning and Visualisation}

The system has been designed to enable users to choose variations of the base MLLM architecture. In the current release, users can choose to fine-tune the following baselines: (i)~NLLB-200-600M, (ii) NLLB-200-1.3M, (iii) NLLB-200-3.3B, or (iv) a user-specified baseline. The fine-tuning mode allow users to specify, using GUI controls, the exact hyperparameters required for the chosen approach.

The visualisation segment provides live graphing of model progression, allowing for the monitoring of model convergence. All log files are preserved and accessible for review to examine the training convergence, as well as to evaluate the model's accuracy during training and validation phases.

\subsection{Deployment}

Gradio (\url{https://gradio.app/}, accessed on 28 November 2023)~\citep{abid2019gradio} is an open-source Python library that enables the development of easy-to-use web applications for machine learning models. The library integrates with the most popular Python libraries, including Scikit-learn and PyTorch. 

A key advantage is that it allows interaction with a web app developed for a Jupyter or Colab notebook. Consequently, it was selected as the library used for the deployment of our custom fine-tuned models. 

\subsection{Green Report}

In recent years, the ecological footprint of technology, along with the assessment of its impacts, has become increasingly prominent~\citep{henderson2020towards}. Indeed, this may be viewed as a natural response to truly massive NLP models which have been developed by large multinational corporations with little apparent regard for their environmental impact.

Specifically, HPO for finely-tuned MLLMs can be especially demanding when the fine-tuning of hyperparameters spans a wide search space.

Consequently, a wide array of tools for assessing NLP's carbon footprint has been created~\citep{bannour-etal-2021-evaluating}, and the idea of sustainable NLP has emerged as a significant area of research. This has been recognised at numerous prestigious conferences; for instance, the Green and Sustainable NLP track at EACL 2021 (\url{https://2021.eacl.org/news/green-and-sustainable-nlp}, accessed on 28 November 2023).

Reflecting these advancements, adaptMLLM has integrated a ``green report'' feature that records the kgCO\textsubscript2 emitted during the development of the model. This aligns closely with the current industry movement towards measuring the environmental impact of NLP activities. 

\subsection{MLLMs: Translation and Evaluation}

Besides facilitating model fine-tuning, the application also provides functionality for translation and assessing model performance. The use of pre-trained models for translation is also parameterised; users specify the model's name as a hyperparameter, which is then used to perform translation and evaluation on the test files.

After building the system, users can select the model they wish to use for translation of the test set. While human judgment is often the most reliable for assessing translation quality, human evaluators are not always accessible, may have differing opinions, and can be costly to engage for experimental purposes. As a result, automatic evaluation metrics are commonly employed, particularly by developers who are tracking the step-by-step advancement of their systems.

Several automatic evaluation metrics provided by SacreBleu (\url{https://github.com/mjpost/sacrebleu}, accessed on 28 November 2023)~\citep{post2018call} are used: BLEU~\citep{papineni-etal-2002-bleu}, TER~\citep{snover2006study} and ChrF~\citep{popovic2015chrf}. Translation quality can also be evaluated using Meteor~\citep{denkowski2014meteor} and F1 score~\citep{melamed-etal-2003-precision}. 

It is important to recognise that BLEU, ChrF, Meteor, and F1 are metrics based on precision, thus higher values signify better performance. On the other hand, TER is a metric based on errors, with lower values denoting superior translation quality. The available evaluation options include standard (truecase) and lowercase BLEU scores, along with sentence-level BLEU scoring, as well as ChrF1 and ChrF3.

Logging occurs at three tiers: a model development log for charting progress, an output log from the training console, and a log of the evaluation outcomes. Additionally, there is a references section that provides materials pertinent to the development, utilisation, and comprehension of adaptMLLM. Presently, validation throughout the training process is performed based on model loss.

\subsection{LLMs: Playgrounds}

When OpenAI (\url{https://openai.com/}, accessed on 28 November 2023) released a playground for its GPT-3 model, the community was quick to create demos. Given that OpenAI's GPT-3 is proprietary, generating text using its API would incorporate a cost and involve sending data to the site. Ideally, we sought to host an open-source text generation model, and associated playground app in our own environment. 

In 2021, Eleuther AI created GPT-J, an open source text generation model to rival GPT-3 and the model is freely available on the Hugging Face Model Hub allowing us to download variations of this model. In this spirit, we have developed our own fully customisable text generation playground using GPT-J. Using Gradio, a web interface that can interact with these GPT-J models was developed. 

\section{Empirical Evaluation}
\label{sec:exp}
After outlining the theoretical framework and the tool itself, we proceed to assess the efficacy of the adaptMLLM methodology by training models for the EN~${\leftrightarrow}$~GA and the EN~${\leftrightarrow}$~MR language pairs.

\subsection{Infrastructure and Hyperparameters}

A Google Colab Pro+ subscription facilitated rapid development of prototypes using NVIDIA 40~GB GPU graphics cards (A100-SXM4-40~GB) and compute resources of up to 89~GB of system memory when available~\citep{Bisong2019}. All MT models were trained using the adaptMLLM application.

The DeepSpeed library (cf. Section \ref{deepspeed}) is a critical component in making the adaptMLLM system work, since it enables our models to be loaded across both GPU and system memory. Without such a library, very significant compute resources would be required which would be prohibitively costly for our team to hire. The hyperparameters used for developing models for both language pairs are outlined in Table \ref{tab:hpo-table}.

\begin{table}[H]
\caption[Hyperparameter Optimisation]{HPO with optimal hyperparameters, within the search space, are highlighted in bold. }
\newcolumntype{L}{>{\raggedright\arraybackslash}X}
\begin{tabularx}{\textwidth}{LL}
\toprule
\textbf{Hyperparameter} & \textbf{Values}      \\ \midrule
Epochs  & 1, 3, \textbf{5} \\ \midrule
Batch size        & 8, 12, \textbf{16}       \\ \midrule
Gradient accumulation steps  & 2,  4, \textbf{8}   \\ \midrule
Learning rate         & 1 $\times~10^{-5}$, \textbf{3 $\mathbold{\times~10^{-5}}$}, 9 $\times~10^{-5}$  \\ \midrule
Weight decay   & 0.01, \textbf{0.1}, 1, 2 \\ \midrule
Mixed precision   &   False, \textbf{True}       \\ \bottomrule
\end{tabularx}

\label{tab:hpo-table}
\end{table}

\subsection{Results: Automatic Evaluation}
To determine the quality of our translations, automated metrics were employed. For comparison with our prior studies, the performance of models was gauged using three evaluative metrics: BLEU, TER, and ChrF. These metrics reflect the precision of translations produced by our finely-tuned MLLM systems. We report case-insensitive BLEU scores at the corpus level. Note that BLEU and ChrF are precision-based metrics, so higher scores are better, whereas TER is an error-based metric and lower scores indicate better translation quality.

\subsubsection{Translation in the EN$~\leftrightarrow~$GA Directions}
The experimental results from the LoResMT2021 Shared Task in the EN$~\leftrightarrow~$GA directions are summarised in Tables \ref{tab:en2ga} and \ref{tab:ga2en} and are compared with our experimental findings, adaptMLLM, achieved by fine-tuning a 3.3B parameter NLLB MLLM. 

The highest-performing EN$~\rightarrow~$GA system in the LoResMT2021 Shared Task was submitted by the ADAPT team~\citep{lankford2021machine}. The model was developed with an in-house application, adaptNMT~\citep{lankford2023adaptNMT} using a Transformer architecture. It performed well across all key translation metrics (BLEU: 36.0, TER: 0.531 and ChrF3: 0.6). 

Subsequently, these results were improved upon (BLEU: 37.6, TER: 0.577 and ChrF3: 0.57) by training a Transformer model on a bespoke health dataset, gaHealth~\citep{lankford2022lrec}.

By fine-tuning the NLLB MLLM, using the parameters outlined in Table \ref{tab:hpo-table}, a significant improvement in translation performance was achieved. The adaptMLLM EN$~\rightarrow~$GA en2ga system, shown in Table \ref{tab:en2ga}, achieves a BLEU score of 41.2, which is 5.2 BLEU points higher than our previous score which won the shared task in 2021. This represents a relative improvement of 14\%.

\begin{table}[H]
\caption[EN~${\rightarrow}$~GA: adaptMLLM systems compared with LoResMT2021]{EN~${\rightarrow}$~GA: adaptMLLM systems compared with LoResMT2021. The impact of fine-tuning the baseline NLLB model is evident with the BLEU score rising from 29.7 to 41.2 representing a 39\% relative improvement. Models developed using adaptMLLM were trained using the optimal hyperparameters set out in Table \ref{tab:hpo-table}.}
\newcolumntype{L}{>{\raggedright\arraybackslash}X}
\begin{tabularx}{\textwidth}{LCCCC}
\toprule
\textbf{Team} &
 \textbf{System} &
  \textbf{BLEU} $\boldsymbol{\uparrow}$ &
  \textbf{TER} $\boldsymbol{\downarrow}$ &
  \textbf{ChrF3} $\boldsymbol{\uparrow}$ \\ \midrule
adaptMLLM & en2ga-tuned & 41.2 & 0.51 & 0.48 \\
adapt & covid\_extended & 36.0 & 0.531 & 0.60 \\
adapt & combined & 32.8 & 0.590 & 0.57 \\
adaptMLLM & en2ga-baseline & 29.7 & 0.595 & 0.559 \\
IIITT  & en2ga-b & 25.8 & 0.629 & 0.53 \\
UCF     & en2ga-b & 13.5 & 0.756 & 0.37   \\
\bottomrule
\end{tabularx}
\label{tab:en2ga}
\end{table}

\vspace{-12pt}
\begin{table}[H]
\caption[GA~${\rightarrow}$~EN: adaptMLLM systems compared with LoResMT2021]{GA~${\rightarrow}$~EN: adaptMLLM systems compared with LoResMT2021. The impact of fine-tuning the baseline NLLB model is evident with the BLEU score rising from 47.8 to 75.1 representing a 57\% relative improvement. Models developed using adaptMLLM were trained using the optimal hyperparameters set out in Table \ref{tab:hpo-table}.}
\label{tab:ga2en}
\newcolumntype{L}{>{\raggedright\arraybackslash}X}
\begin{tabularx}{\textwidth}{LCCCC}
\toprule
\textbf{Team} &
 \textbf{System} &
  \textbf{BLEU} $\boldsymbol{\uparrow}$ &
  \textbf{TER} $\boldsymbol{\downarrow}$ &
  \textbf{ChrF3} $\boldsymbol{\uparrow}$ \\ \midrule
adaptMLLM & ga2en-tuned & 75.1 & 0.385 & 0.71\\
adaptMLLM & ga2en-baseline & 47.8 & 0.442 & 0.692 \\
IIITT  & ga2en-b & 34.6 & 0.586 & 0.61\\
UCF & ga2en-b & 21.3 & 0.711 & 0.45\\
\bottomrule
\end{tabularx} 
\end{table}

For translation in the GA$~\rightarrow~$EN direction, illustrated in Table \ref{tab:ga2en}, the best-performing model for the LoResMT2021 Shared Task was developed by IIITT with a BLEU of 34.6, a TER of 0.586 and ChrF3 of 0.6. Accordingly, this serves as the baseline score by which we can benchmark our GA$~\rightarrow~$EN model, developed by fine-tuning a 3.3B parameter NLLB using adaptMLLM. Similar to the results achieved in the EN$~\rightarrow~$GA direction, significant improvement in translation performance was observed using this new method. The performance of the adaptMLLM model offers an improvement across all metrics with a BLEU score of 75.1, a TER of 0.385 and a ChrF3 result of 0.71. In particular, the 117\% relative improvement in BLEU score against the IIITT system is very significant. The adaptMLLM model is a fine-tuned pre-trained NLLB 3.3B parameter MLLM, whereas the IIITT model fine-tuned a smaller Opus MT model from Helsinki NLP. MLLMs and LLMs have already learned to represent natural language patterns and structures from large amounts of data, which can be adapted to specific tasks or domains by updating the model's parameters with a smaller amount of annotated data. The effect of this approach is demonstrated in the substantially higher BLEU achieved by the adaptMLLM model relative to the IIITT model which was trained on a much smaller Opus model.

The improvement in translation performance is real and not just a BLEU score anomaly given that large improvements were simultaneously observed across the BLEU, TER and CHRF metrics. More specifically, Meta's nllb-200-3.3B model has a memory footprint of 17.58 GB enabling 3.3 billion parameters to be trained compared to the Helsinki-NLP model, opus-mt-ga-en, which is just 295 MB and has a correspondingly much smaller set of trainable parameters. Another aspect differentiating the adaptMLLM approach is the relatively broad hyperparameter search space compared to systems developed by other teams which are outlined in Table \ref{tab:ga2en}. We experimented with the number of epochs, the batch size, the gradient accumulation steps, the learning rate, the weight decay and the type of precision used. The exact hyperparameters used are illustrated in Table \ref{tab:hpo-table}.

\subsubsection{Translation in the EN$~\leftrightarrow~$MR Directions}

The experimental results from the LoResMT2021 Shared Task in the EN$~\leftrightarrow~$MR directions are summarised in Tables \ref{tab:en2mr} and \ref{tab:mr2en}, and are compared with our experimental findings in developing adaptMLLM. For the shared task, the highest-performing EN$~\rightarrow~$MR system was submitted by the IIITT team. Their model used a Transformer architecture and achieved a BLEU score of 34.6, a TER of 0.586, and ChrF3 of 0.61. 

\begin{table}[H]
\caption[EN~${\rightarrow}$~MR: adaptMLLM systems compared with LoResMT2021]{EN~${\rightarrow}$~MR: adaptMLLM systems compared with LoResMT2021. The impact of fine-tuning the baseline NLLB model is evident with the BLEU score rising from 19.8 to 26.4, representing a 33\% relative improvement. Models developed using adaptMLLM were trained using the optimal parameters set out in Table \ref{tab:hpo-table}.}
\newcolumntype{L}{>{\raggedright\arraybackslash}X}
\begin{tabularx}{\textwidth}{Lm{3cm}<{\centering}CCC}
\toprule
\textbf{Team} &
 \textbf{System} &
  \textbf{BLEU} $\boldsymbol{\uparrow}$ &
  \textbf{TER} $\boldsymbol{\downarrow}$ &
  \textbf{ChrF3} $\boldsymbol{\uparrow}$ \\ \midrule
adaptMLLM & en2mr-tuned & 26.4 & 0.56 & 0.608 \\  
IIITT & en2mr-IndicTrans-b & 24.2 & 0.59 & 0.597   \\
oneNLP-IIITH & en2mr-Method2-c & 22.2 & 0.56 & 0.746 \\
oneNLP-IIITH & en2mr-Method3-c & 22.0 & 0.56 & 0.753 \\
oneNLP-IIITH & en2mr-Method1-c & 21.5 & 0.56 & 0.746 \\
adaptMLLM & en2mr-baseline & 19.8 & 0.656 & 0.57 \\
adaptNMT & en2mr & 13.7 & 0.778 & 0.393\\
\bottomrule
\end{tabularx}

\label{tab:en2mr}
\end{table}

\vspace{-12pt}
\begin{table}[H]
\caption[MR~${\rightarrow}$~EN: adaptMLLM systems compared with LoResMT2021]{MR~${\rightarrow}$~EN: adaptMLLM systems compared with LoResMT2021. The impact of fine-tuning the baseline NLLB model is evident with the BLEU score rising from 42.7 to 52.6, representing a 23\% relative improvement. Models developed using adaptMLLM were trained using the optimal hyperparameters set out in Table \ref{tab:hpo-table}.} 
\newcolumntype{L}{>{\raggedright\arraybackslash}X}
\begin{tabularx}{\textwidth}{m{2.5cm}<{\raggedright}m{5cm}<{\centering}CCC}
\toprule
\textbf{Team} &
 \textbf{System} &
  \textbf{BLEU} $\boldsymbol{\uparrow}$ &
  \textbf{TER} $\boldsymbol{\downarrow}$ &
  \textbf{ChrF3} $\boldsymbol{\uparrow}$ \\ \midrule
adaptMLLM & mr2en-tuned & 52.6 & 0.409 & 0.704\\  
adaptMLLM & mr2en-baseline & 42.7 & 0.506 & 0.639 \\
oneNLP-IIITH & mr2en-Method3-c & 31.3 & 0.58 & 0.646 \\
oneNLP-IIITH & mr2en-Method2-c & 30.6 & 0.57 & 0.659 \\
oneNLP-IIITH & mr2en-Method1-c & 20.7 & 0.48 & 0.735 \\
adaptNMT & mr2en & 19.9 & 0.758 & 0.429\\
UCF & mr2en-UnigramSegmentation-b & 7.7 & 0.24 & 0.833 \\
IIITT & mr2en-IndicTrans-b & 5.1 & 0.22 & 1.002 \\
\bottomrule
\end{tabularx} 

\label{tab:mr2en}
\end{table}

Again the approach taken by adaptMLLM in fine-tuning a 3.3.B parameter NLLB MLLM yielded the best performance  compared with other systems entered for the shared task. The EN$~\rightarrow~$MR adaptMLLM en2mr system achieves the highest BLEU score of 26.4 compared with IIITT, the winning team in the EN$~\rightarrow~$MR shared task. IIITT had a BLEU score of 24.2 which represents a relative improvement of 9\% for the adaptMLLM system. The other key translation metrics of TER and ChrF3 were also improved upon indicating that the adaptMLLM system is the best approach in the EN$~\rightarrow~$MR direction.  

\par
For translation in the MR~${\rightarrow}$~EN direction, the best-performing model for the \sloppy{LoResMT2021} Shared Task was developed by oneNLP-IIITT with a BLEU score of 31.3, a TER of 0.58 and ChrF3 of 0.646. This serves as the baseline score by which our MR~${\rightarrow}$~EN model, developed using adaptMLLM, can be benchmarked. The performance of the adaptMLLM model offers a significant improvement across all metrics with a BLEU score of 52.6, a TER of 0.409 and a ChrF3 of 0.704. Again this represents a very strong relative improvement of 68\% in BLEU compared with the winning team from the shared task.

\subsection{Human Evaluation Results}

Irish, characterised by its complex morphology, flexible sentence structure, and extensive inflection, presents unique challenges in translation from English. As a result, accurately producing grammatical aspects like gender or case inflections in nouns within Irish translations often proves to be a difficult task.

\par
This research aims to investigate the manner in which a neural machine translation (NMT) system, like a fine-tuned NLLB model, manages these linguistic complexities. Current studies imply that fine-tuned MLLMs are likely to enhance these language features~\citep{costa2022no}. MLLMs and LLMs tackle the issue indirectly through subword models in an unsupervised fashion, without grasping the explicit formal principles of grammatical categories.

\par
Past human evaluation studies examining EN$~\rightarrow~$GA MT performance have centred on outputs from NMT systems that did not use pre-trained models~\citep{lankford2022human}. In the context of this research, we now conduct human evaluation on the output from our MLLM models. The work is further differentiated in that it examines the output in both the EN$~\rightarrow~$GA and GA$~\rightarrow~$EN directions. The approach taken in the previous study and our current work are similar in that we use SQM and MQM as our human evaluation metrics.
\par
While automatic evaluation metrics show that a fine-tuned MLLM approach leads to significant improvements compared to building a Transformer model from scratch, it fails to address the issue of grammatical or linguistic quality in the translated output. Such an approach does not account for the subtleties of handling gender or cases in the target language. To gain a more comprehensive understanding of the linguistic errors produced by MLLM systems, a fine-grained human evaluation was conducted through a manual error analysis. This approach allowed for the identification and categorisation of specific translation errors associated with each of the evaluated systems, providing a foundation for future work aimed at improving the translation quality of the models.

\par 
We also describe the annotation framework, the overall annotation process, and the level of agreement among annotators, which broadly follows the approach taken by other fine-grained human evaluation studies~\citep{klubivcka2018quantitative,lankford2022human}.

\subsubsection{Scalar Quality Metrics}

The SQM framework modifies the WMT shared-task settings to acquire segment-level scalar ratings with document context. SQM assesses the quality of translations using a scale that ranges from 0 to 6, which is different from the WMT approach~\citep{ma-etal-2017-blend}, which employs a range of 0 to 100.

When using this evaluation method, annotators are required to choose a rating ranging from 0 to 6 after being presented with the source and target sentences. Table \ref{tab:sqm} provides the SQM quality levels for ratings 0, 2, 4, and 6. In situations where the translations do not precisely align with the core SQM levels, annotators may select intermediate ratings of 1, 3, or 5.

\begin{table}[H]
\caption{SQM levels explained~\citep{freitag2021experts}.}								
\begin{tabularx}{\textwidth}{Cm{11cm}<{\raggedright}}
\toprule
\multicolumn{1}{c}{\textbf{SQM Level}} &
  \textbf{Details of Quality} \\ \midrule
6 &
  Perfect Meaning and Grammar: The meaning of the translation is completely consistent with the source and the surrounding context (if applicable). The grammar is also correct. \\ \midrule
4 &
  Most Meaning Preserved and Few Grammar Mistakes: The translation retains most of the meaning of the source. This may contain some grammar mistakes or minor contextual inconsistencies. \\  \midrule
2 & Some Meaning Preserved: The translation preserves some of the meaning of the source but misses significant parts. The narrative is hard to follow due to fundamental errors. Grammar may be poor. \\  \midrule
0 &
  Nonsense/No meaning preserved: Nearly all information is lost between the translation and source. Grammar is irrelevant. \\ \bottomrule
\end{tabularx}
\label{tab:sqm}

\end{table}

\par
The average annotator SQM scores arising from our human evaluation were compared with automatic metric scores recorded by adpatMLLM when evaluating the EN~${\leftrightarrow}$~GA systems. These results, illustrated in Table \ref{tab:sqm_results}, indicate a high level of correlation between the automatic metrics and the SQM outputs of the human evaluation. Clearly, the system translating in the GA~${\rightarrow}$~EN direction performs better, when evaluated using both automatic and human evaluation, than its counterpart when translating in the opposite direction. These results are consistent with our previous work, which also show better GA~${\rightarrow}$~EN translation performance~\citep{lankford2023adaptNMT}. This performance difference is  attributed to the morphologically rich nature of the Irish language, which relies heavily on inflection, derivation, and its case system. 
\begin{table}[H]
\caption[Annotator SQM scores for adaptMLLM systems]{Average SQM scores for adaptMLLM systems compared with automatic metrics.}
\newcolumntype{L}{>{\raggedright\arraybackslash}X}
\begin{tabularx}{\textwidth}{m{3cm}<{\raggedright}CCCC}
\toprule
\textbf{System} &
  \textbf{BLEU} $\boldsymbol{\uparrow}$ &
  \textbf{TER} $\boldsymbol{\downarrow}$ &
  \textbf{ChrF3} $\boldsymbol{\uparrow}$ &
  \textbf{SQM} $\boldsymbol{\uparrow}$ \\ \midrule
adaptMLLM en2ga    & 41.2  & 0.51  & 0.48 & 4.38  \\
adaptMLLM ga2en   & 75.1  & 0.385 & 0.71 & 5.63 \\
\bottomrule
\end{tabularx}
\label{tab:sqm_results}
\end{table}

\subsubsection{Multidimensional Quality Metrics}

Within the QTLaunchpad project (\url{https://www.qt21.eu}, accessed on 28 November 2023), the development of~the MQM framework (\url{https://www.qt21.eu/mqm-definition/definition-2015-12-30.html}, accessed on 28 November 2023) aimed to offer a structured approach to conducting manual evaluations through meticulous error analysis. This framework does not mandate a uniform metric for all applications; rather, it supplies an extensive list of potential quality issues, each with standardised names and definitions, which can be tailored to particular tasks. Beyond establishing a dependable method for quality evaluation, the MQM framework also enables us to identify and select error tags pertinent to our specific task.

\par
We customised the MQM framework to suit our context by following the official scientific research guidelines~\citep{Lommel2018}. Our modifications to MQM are explained below.

\par
The original MQM guidelines propose a wide range of tags on different annotation layers. However, for our specific annotation task, this comprehensive tagset is too detailed. Hence, we evaluated our MT output using the smaller default set of evaluation categories outlined in the core tagset. These standard top-level categories, which include accuracy and fluency, are recommended by the MQM guidelines and are presented in Table \ref{tab:mqmcat}.

\begin{table}[H]
\caption[Description of error categories within the core MQM framework]{Description of error categories within the core MQM framework~\citep{freitag2021experts}.}
\newcolumntype{L}{>{\raggedright\arraybackslash}X}
\begin{adjustwidth}{-\extralength}{0cm}
\begin{tabularx}{\fulllength}{LLm{10cm}<{\raggedright}}
\toprule
\textbf{Category} & \textbf{Sub-Category} & \textbf{Description}                                 \\ \midrule
Non-translation 
 &                & Impossible to reliably characterise the 5 most severe errors.       \\ 
       \midrule
Accuracy & Addition       & Translation includes information not present in the source.         \\
         & Omission              & Translation is missing content from the source.      \\
                & Mistranslation & Translation does not accurately represent the source.               \\
         & Untranslated text     & Source text has been left untranslated.              \\
        \midrule
Fluency & Punctuation           & Incorrect punctuation                                \\
         & Spelling              & Incorrect spelling or capitalisation.                \\
         & Grammar               & Problems with grammar, other than orthography.       \\
                         & Register       & Wrong grammatical register (e.g., inappropriately informal pronouns). \\
                  & Inconsistency         & Internal inconsistency (not related to terminology). \\
                  & Character encoding    & Characters are garbled due to incorrect encoding.   \\ \bottomrule
\end{tabularx}
\end{adjustwidth}
\label{tab:mqmcat}
\end{table}

We used a special non-translation error tag to label entire sentences that were so poorly translated that individual errors could not be identified. Error severities were designated as major or minor errors, and they were assigned independently of the category. These corresponded to actual translation or grammatical errors and minor imperfections, respectively. We used the default recommended weights~\citep{Lommel2018}, which assign a weight of 1 to minor errors, while major errors are given a weight of 10. Additionally, the non-translation category was assigned a weight of 25, which is consistent with best practice established in previous studies~\citep{freitag2021experts}.

Our annotators were instructed to identify all errors in each sentence of the translated output using the error categories provided in Table \ref{tab:mqmcat}.
\subsubsection{Annotation~Setup}

Annotations were carried out using a detailed, fine-grained MQM approach and a simpler SQM approach. The SQM categories are summarised in Table \ref{tab:sqm} whereas the hierarchical taxonomy of our MQM implementation is outlined in Table \ref{tab:mqmcat}.

\par
Working independently of one another, two annotators with similar backgrounds were selected for the annotation of fine-tuned EN$~\leftrightarrow~$GA systems. Both annotators are fluent speakers of Irish and neither had prior experience with MQM. The annotators are postgraduate students of the Máistir Gairmiúil san Oideas (Postgraduate Masters in Education) at the University of Galway (\url{https://universityofgalway.ie}, accessed on 28 November 2023).

Before starting the annotation process, they were extensively briefed on the process and the MQM annotation guidelines. These guidelines provide in-depth directions for carrying out annotation activities under the MQM framework.

\par
In conducting the EN$~\rightarrow~$GA human evaluation of the translation output, we presented our annotators with a test set of 25 randomly selected sentences, which consisted of the English source text, an Irish reference translation and the unannotated fine-tuned MLLM EN$~\rightarrow~$GA system output. 

A similar approach was adopted for the GA$~\rightarrow~$EN human evaluation where the annotator test set consisted of 25 randomly selected sentences, which consisted of the Irish source text, an English reference translation and the unannotated fine-tuned MLLM GA$~\rightarrow~$EN system output.

\par
After extracting the annotation data, the annotators individually examined the output to assess the performance of each system across the different error categories. 
 
\subsubsection{Inter-Annotator Agreement}

In order to ensure the validity of our research findings, it is essential to assess the degree of consensus among our annotators~\citep{artstein2017inter}. Manual evaluation methods for MT, such as MQM, often result in low inter-annotator agreement (IAA)~\citep{lommel-etal-2014-using,callison-burch-etal-2007-meta}. We computed inter-annotator agreement using Cohen's kappa ($k$) coefficient~\citep{cohen1960coefficient}, a widely recognised metric in the field. The evaluation was performed at the sentence level for each individual system, and the agreement discrepancies across systems were examined. This approach also allowed us to obtain an overall view of the level of agreement between annotators.

Table \ref{tab:mqmtotals-human} highlights the cumulative number of errors identified by the annotators for each system. Looking at the aggregate data alone, it is evident that both annotators have judged the EN$~\rightarrow~$GA system to contain significantly more errors, which supports the findings of the automatic evaluation. 

\begin{table}[H]
\caption{System errors found by each annotator using the MQM metric.}
\newcolumntype{L}{>{\raggedright\arraybackslash}X}
\begin{tabularx}{\textwidth}{m{5cm}<{\raggedright}CC}
\toprule 
\textbf{Num Errors}       & \textbf{EN$~\boldsymbol{\rightarrow}~$GA}                 & \textbf{GA$~\boldsymbol{\rightarrow}~$EN}                \\ \midrule
Annotator 1  & 53                     & 7       \\ \midrule
Annotator 2  & 82                     & 11       \\ \bottomrule
\end{tabularx}
\label{tab:mqmtotals-human}
\end{table}

\par
Table \ref{tab:mqmtotals-human} provides a useful overview for evaluating which system performs better overall, but it does not offer the detailed analysis necessary to identify specific linguistic areas for improvement in the translations. For a more comprehensive understanding, we delved into a detailed examination of the types of errors present, with the findings presented in Table \ref{tab:combined}. This table breaks down the total number of error tags noted by each annotator for each system, categorised by the type of error. The detailed analysis underscores how the GA$~\rightarrow~$EN system outperforms the EN$~\rightarrow~$GA system. Notably, the GA$~\rightarrow~$EN system's translations display significantly greater fluency, as evidenced by just two errors recorded in this category.

\par 
One way to measure inter-rater reliability is to use Cohen's kappa, which is a rigorous method. It determines the percentage of items that raters agree on while also taking into account the possibility of them agreeing on some items by chance. Cohen's kappa was calculated separately for every error type and the findings are outlined in Table \ref{tab:my-cohen} and discussed in further detail later in Section \ref{obs}. To calculate Cohen's kappa the following formula is used: 
\begin{equation} \label{e1}
k = (p\textsubscript{o} - p\textsubscript{e}) / (1 - p\textsubscript{e}) 
\end{equation}
\begin{enumerate}[label=,leftmargin=*,labelsep=0mm]
\item[ ] $p\textsubscript{o}$: Relative observed agreement among raters
\item[ ] $p\textsubscript{e}$: Hypothetical probability of chance agreement.
\end{enumerate}

\begin{table}[H]

\caption{Fine-grained analysis with concatenated errors across both annotators.}
\newcolumntype{L}{>{\raggedright\arraybackslash}X}
\begin{tabularx}{\textwidth}{m{5cm}<{\raggedright}CC}
\toprule 
\textbf{Error Type} &
  \multicolumn{1}{c}{\textbf{EN$~\boldsymbol{\rightarrow}~$GA Errors}} & \textbf{GA$~\boldsymbol{\rightarrow}~$EN Errors} \\ \midrule
Non-translation     & 0 & 0  \\
Accuracy 
&  \\
\quad Addition                                  & 12 & 5 \\
\quad Omission                                  & 14 & 3 \\
\quad Mistranslation                            & 41 & 6 \\
\quad Untranslated text                         & 9 & 2 \\
Fluency               &  \\
\quad Punctuation                               & 10 & 0 \\
\quad Spelling                                  & 6 & 0 \\
\quad Grammar                                   & 27 & 0 \\
\quad Register                                  & 19 & 2 \\
\quad Inconsistency                             & 6 & 0 \\
\quad Character Encoding                        & 0 & 0\\ \midrule
Total errors & 135 & 18 \\ \bottomrule
\end{tabularx}
\label{tab:combined}
\end{table}

\subsubsection{Inter-Annotator Reliability}\label{sec:rater}

In Cohen's seminal paper~\citep{cohen1960coefficient}, he precisely defines the interpretation of various $k$ scores. Scores $\le$ 0 indicate no agreement, scores from 0.01 to 0.20 suggest none to slight agreement, scores from 0.21 to 0.40 denote fair agreement, scores from 0.41 to 0.60 reflect moderate agreement, scores from 0.61 to 0.80 correspond to substantial agreement, and scores from 0.81 to 1.00 represent almost perfect agreement. 
The kappa values of each error type are displayed in Table \ref{tab:my-cohen}.
\begin{table}[H]
\centering
\caption[Inter-annotator agreement using Cohen values]{Inter-annotator agreement using Cohen values. Perfect observed agreement is indicated by $p\textsubscript{o}$ = 1.}
\newcolumntype{L}{>{\raggedright\arraybackslash}X}
\begin{tabularx}{\textwidth}{m{5cm}<{\centering}CC}
\toprule 
\textbf{Error Type} & \multicolumn{1}{c}{\textbf{EN$~\boldsymbol{\rightarrow}~$GA}} & \textbf{GA$~\boldsymbol{\rightarrow}~$EN}  \\ \midrule
\multicolumn{1}{l}{Non-translation}               & $p\textsubscript{a}$~=~1 & $p\textsubscript{a}$~=~1 \\
\multicolumn{1}{l}{Accuracy}                      &  &  \\
Addition                                          & 0.24 & 0 \\
Omission                                          & 0.31  & 0 \\
Mistranslation                                    & 0.32 & $-$0.11 \\
Untranslated text                                 & 0.07 & 0 \\
\multicolumn{1}{l}{Fluency}                       &  &  \\
Punctuation                                       & 1 & $p\textsubscript{o}$~=~1 \\
Spelling                                          & 0.24 & $p\textsubscript{o}$~=~1 \\
Grammar                                           & 0.59 & $p\textsubscript{o}$~=~1 \\
Register                                          & $-$0.07 & 0 \\
Inconsistency                                     & 0.34 & $p\textsubscript{o}$~=~1 \\
Character Encoding                                & $p\textsubscript{o}$~=~1 & 1.0\\ 
 \bottomrule
\end{tabularx}

\label{tab:my-cohen}
\end{table}  

Many chance-adjusted indices of inter-rater reliability estimate agreement using a distribution-based approach. A problem arises when there is only one observed response category, resulting in a score of NaN (Not a Number). This occurs when the observed agreement, $p\textsubscript{o}$ and the chance agreement, $p\textsubscript{e}$ are both 1, which cannot be computed as seen in Equation (\ref{e1}). In such cases, it is better to report $p\textsubscript{o}$ instead of $kappa$, since there is perfect observed agreement, i.e., $p\textsubscript{o}$ = 1.

As illustrated in Table \ref{tab:my-cohen}, we observe a high level of agreement overall. There is either fair agreement, or perfect observed agreement, in 16 out of 22 sub-categories. Given these scores, we have a high degree of confidence in the human evaluation of the fine-tuned MLLM outputs.
 
\subsection{Environmental Impact}
\label{sec:envimp}

Motivated by research which examines the environmental impact of NLP~\citep{strubell-etal-2019-energy,10.1145/3442188.3445922}, we monitored the energy and carbon emissions required to train our models.

Model development was carried out using Colab Pro+, which as part of Google Cloud is carbon neutral~\citep{lacoste2019quantifying}. All fine-tuning experiments of MLLMs were conducted on Google Cloud servers and consequently were emission free (\url{https://cloud.google.com/sustainability/region-carbon}, accessed on 28 November 2023).

In terms of energy consumption, the total power draw for each experimental run is outlined in Table \ref{tab:environ}. As part of our Google Colab subscription, Nvidia a100-sxm4-40gb graphics cards were used which have a max power consumption of 400~W. The calculations are based on the graphics card running at 80\% max power during model training.  

\begin{table}[H]

\caption[\hl{Energy} consumption during MLLM fine-tuning experiments]{Energy consumption during MLLM fine-tuning experiments. All experiments carried out on Google Cloud with 0 kgCO\textsubscript{2} emissions.}
\newcolumntype{L}{>{\raggedright\arraybackslash}X}
\begin{tabularx}{\textwidth}{m{3cm}<{\raggedright}LLLLLL}
\toprule 
\textbf{System} &
  \textbf{BLEU} $\boldsymbol{\uparrow}$ &
  \textbf{TER} $\boldsymbol{\downarrow}$ &
  \textbf{ChrF3} $\boldsymbol{\uparrow}$ &
  \textbf{Lines} &
  \textbf{Runtime (Hours)} &
  \textbf{kWh} \\ \midrule
adaptMLLM en2ga    & 41.2  & 0.51  & 0.48 & 13~k  & 3.51 & 1.1  \\
adaptMLLM ga2en   & 75.1  & 0.385 & 0.71 & 13~k  & 3.41 & 1.1  \\
adaptMLLM en2mr  & 26.4 & 0.56 & 0.608 & 21~k & 5.49 & 1.8  \\
adaptMLLM mr2en  & 52.6 & 0.409 & 0.74 & 21~k & 5.43 & 1.7 \\ 
\bottomrule
\end{tabularx}
\label{tab:environ}
\end{table}

\section{Discussion} \label{sec:discussion}
 
We used the adaptMLLM application to create MT models with datasets from the LoResMT2021 Shared Task in order to assess system efficiency when translating in the EN$~\leftrightarrow~$GA directions.

High-performing models achieving SOTA scores were developed by fine-tuning the NLLB MLLM pretained models with adaptMLLM. Using an easily-understood framework such as adaptMLLM, the benefits of developing high-performing fine-tuned models with small in-domain datasets is thus clear.

\subsection{Performance of adaptMLLM Models Relative to Google Translate}

Translation engine performance, at the corpus level, was benchmarked against Google Translate's (\url{https://translate.google.com}, accessed on 28 November 2023) EN~${\leftrightarrow}$~GA translation service, which is freely available on the internet. 

A full evaluation of Google Translate's engines on the EN$~\rightarrow~$GA test set generated a BLEU score of 38.7, a TER score of 0.493 and a ChrF3 of 0.633. The comparative scores on the test set using our fine-tuned MLLM realised 41.2 for BLEU, 0.489 for TER and 0.653 for ChrF3. Therefore, in the EN$~\rightarrow~$GA direction, the adaptMLLM system demonstrates a relative BLEU score improvement of 6.5\% compared to Google Translate.

The translation output from our fine-tuned MLLMs was also compared with Google Translate using random samples from the LoResMT2021 EN$~\rightarrow~$GA corpus. Table \ref{tab:translations_en2ga} highlights random samples which were picked from the English source test file. A perfect match, with a BLEU of 100, was recorded in one instance, which is unusual. However, this may occur on occasion with the translation of short sentences. Any duplicates between training and test data were removed prior to fine-tuning, but the possibility exists of the test sentence forming part of the original training of the NLLB model exists.

Translation of these samples was independently carried out on the optimal fine-tuned MLLM model and also using Google Translate. Case-insensitive, sentence-level BLEU scores were recorded and are presented in Table \ref{tab:en2ga_gt}.

\begin{table}[H]
\caption{EN~${\rightarrow}$~GA test dataset of LoResMT2021: samples of human reference translations.}
\newcolumntype{L}{>{\raggedright\arraybackslash}X}
\begin{tabularx}{\textwidth}{LL}
\toprule 
\textbf{Source Language (English)} & \textbf{Human Translation (Irish)} \\ \midrule
Temporary COVID-19 Wage Subsidy Scheme & Scéim Fóirdheontais Shealadaigh Pá COVID-19\\\midrule
how COVID-19 spreads and its symptoms & conas a scaipeann COVID-19 agus na siomptóim a bhaineann leis\\\bottomrule
\end{tabularx}
\label{tab:translations_en2ga}
\end{table}

\vspace{-12pt}
\begin{table}[H]
\caption{EN~${\rightarrow}$~GA fine-tuned MLLM model compared with Google Translate. }
\newcolumntype{L}{>{\raggedright\arraybackslash}X}

\begin{tabularx}{\textwidth}{Lm{2cm}<{\raggedright}Lm{2cm}<{\raggedright}}
\toprule 
\textbf{Fine-Tuned LLM} & \textbf{BLEU} {$\boldsymbol{\uparrow}$} & \textbf{Google Translate} & \textbf{BLEU} {$\boldsymbol{\uparrow}$}\\ \midrule
Scéim Fóirdheontais Pá Sealadach COVID-19 & 25.4 & Scéim Fóirdheontais Pá Shealadach COVID-19 & 25.4 \\\midrule
Conas a scaipeann COVID-19 agus na comharthaí a bhaineann leis & 100 & conas a scaipeann COVID-19 agus na hairíonna a bhaineann leis & 65.8 \\\bottomrule
\end{tabularx}
\label{tab:en2ga_gt}
\end{table}

The translation output from our fine-tuned MLLMs was also compared with Google Translate using random samples from the LoResMT2021 EN$~\rightarrow~$MR corpus. A full evaluation of Google Translate's engines on the EN$~\rightarrow~$MR test set, with 500 lines, generated a BLEU score of 25.9, a TER score of 0.566 and a a ChrF3 of 0.601. The comparative scores on the test set using our fine-tuned MLLM realised 26.4 for BLEU, 0.565 for TER, and 0.608 for ChrF3. Therefore, in the EN$~\rightarrow~$MR direction, the adaptMLLM system demonstrates a relative BLEU score improvement of 1.9\% compared to Google Translate.

Samples from the EN$~\rightarrow~$MR test set, along with the corresponding human translation, are illustrated in Table \ref{tab:translations_en2mr}. The performance of these individual samples from the MLLM output and the Google Translation output is compared in Table \ref{tab:en2mr_gt}. The results are promising and suggest that our translation models perform well on the datasets from LoResMT2021. 

\begin{table}[h]
\caption{Samples of human reference translations from EN${\rightarrow}$MR LoResMT2021}
\includegraphics[width=1.0\textwidth]{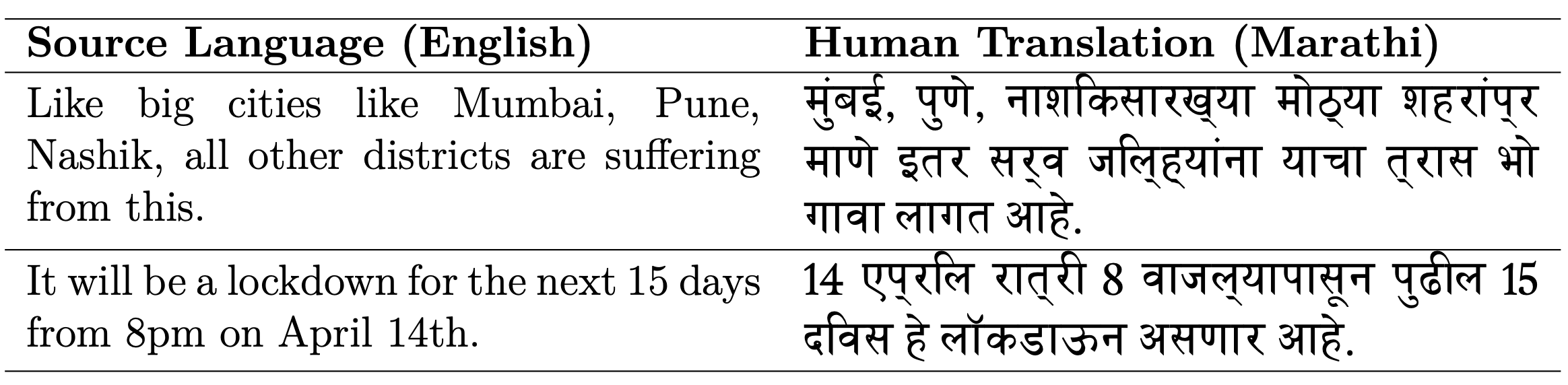}
\label{tab:translations_en2mr}
\end{table}

\begin{table}[h]
\caption[EN${\rightarrow}$MR fine-tuned MLLM model compared with Google Translate]{EN${\rightarrow}$MR fine-tuned MLLM model compared with Google Translate. MR phrases are back-translated to EN and highlighted immediately below each MR sentence pair.}
\includegraphics[width=1.0\textwidth]{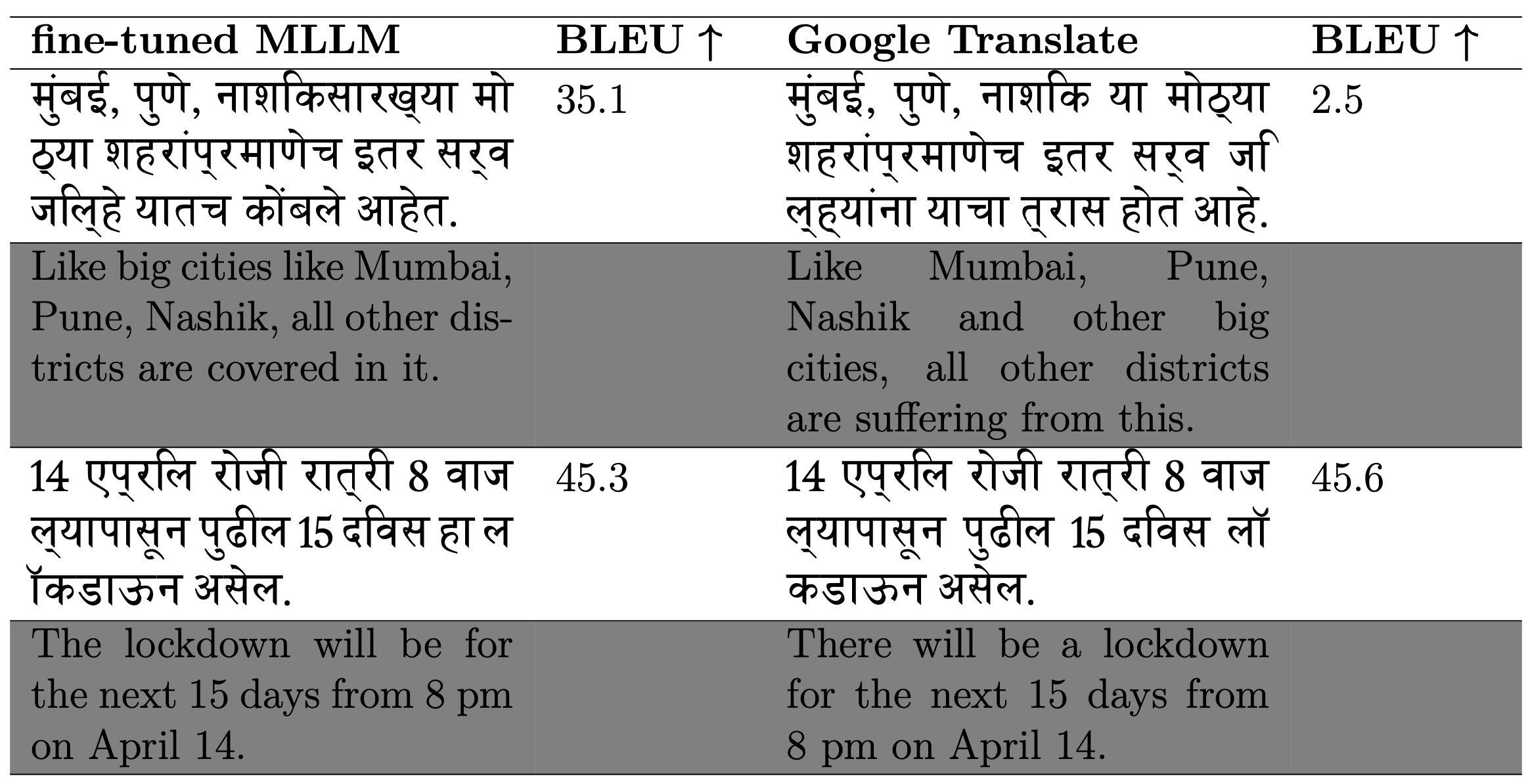}
\label{tab:en2mr_gt}
\end{table}

\subsection{Linguistic Observations}\label{obs}

Table \ref{tab:ling-anal} provides a linguistic analysis of the EN$~\rightarrow~$GA MLLM outputs, showcasing the source sentences alongside their corresponding translations. These sentences were chosen specifically for this detailed human evaluation since they underscore the principal types of errors observed. The approach adopted is similar to the analysis taken in our previous human evaluation of EN$~\rightarrow~$GA translation~\citep{lankford2022human}, in that it focuses on model output errors which fall into the categories: `interpreting meaning' and `core grammatical errors'.  

\begin{table}[H]

\caption{Linguistic analysis of EN$~\rightarrow~$GA system output. Errors in the target translation are flagged in red and the corresponding original source is highlighted in blue. 
}
\begin{tabularx}{\textwidth}{Lm{11cm}<{\raggedright}}
\toprule
\textbf{Type} &
  \textbf{Sentence} \\ \midrule
EN-1 & 
\textcolor{blue}{COVID-19 information} and advice for taxpayers and agents \\ \midrule
GA-1 &
Eolas agus \textcolor{red}{comhairle COVID-19} díocóirí cánach agus dionadaithe \\ \midrule
EN-2 &
 \textcolor{blue}{We understand} the unprecedented situation facing taxpayers as a result of the COVID-19 pandemic.    \\ \midrule
GA-2 &
\textcolor{red}{Tuigeann muid} an cás gan fasach atá roimh cháiníocóirí mar thoradh ar an bpaindéim COVID-19. \\ \midrule
EN-3 &
Further information on Employment Wage Subsidy Scheme (EWSS) is available from the Employing people \textcolor{blue}{section} on this website. \\\midrule
GA-3 &
Tá tuilleadh faisnéise ar Scéim Fóirdheontais Pá Fostaíochta (EWSS) ar fáil ón gcuid Fostaithe ar an \textcolor{red}{láithreán} gréasáin seo.\\ \midrule
EN-4 &
Information for \textcolor{blue}{employers} on the Temporary COVID-19 Wage Subsidy Scheme is available from the Employing people section on this website. \\ \midrule
GA-4 &
Tá faisnéis \textcolor{red}{dfhostóirí} ar an Scéim Fóirdheontais Pá Sealadach COVID-19 ar fáil ón gcuid Fostaithe ar an láithreán gréasáin seo.\\ \bottomrule

\end{tabularx}

\label{tab:ling-anal}
\end{table}

\subsubsection{Interpreting Meaning}

When examining the relationship of one noun to another noun, it should not necessarily be directly translated from English to Irish. This is illustrated in EN-2, where “COVID-19 information and advice” refers to the information and advice that is related to COVID. However, the ENGA system translates this to “Comhairle COVID-19”, which effectively means “COVID-19's information and advice”, i.e., COVID-19 is treated as a possessive noun, which is incorrect.

At times the translated output does not reflect the context in which particular words should be used. An example of this can be seen in the translation of the word “Employer's section” in EN-3, which was interpreted by the ENGA system as “gcuid Fostaithe”. In this English source sentence, the meaning focuses on a section related to a website and the correct translation would be “rannán Daoine a Fhostú”. This is outlined in more detail on the reference website, Fóclóir (\url{https://www.focloir.ie}, accessed on 28 November 2023). It is interested to note that Google Translate correctly interprets this meaning in its translation of the sentence.

Given the nature of the source text, one word frequently encountered was ``Information''. The word was accurately translated to ``faisnéis'' over the text, but it is important to note this word is not widely used in the Irish language. We recommend using the word ``eolas'' (knowledge), since it is a more natural and intuitive translation (\url{https://www.teanglann.ie/en/fgb/eolas}, accessed on 28 November 2023).

\subsubsection{Core Grammatical Errors}

Common mistakes which were encountered throughout the texts involved the use of the apostrophe. Most of these mistakes were flagged as minor errors, but in some cases a missing apostrophe conveyed an entirely different meaning. An example of this can be seen in EN-4 and GA-4 where “information for employers” has been translated to  “faisnéis dfhostóirí” which means “employers' information”. By simply correcting this to “faisnéis d'fhostóirí”, the correct meaning would have been preserved.

\section{Conclusions and Future Work}\label{concl}

We presented adaptMLLM, a comprehensive application designed for the fine-tuning of MLLMs that handles the entire process of model development, evaluation, and deployment. The performance of the application was showcased through the creation of EN$~\leftrightarrow~$GA translation models, which exhibited substantial improvements over the top-ranked models from the EN$~\leftrightarrow~$GA LoResMT2021 Shared Tasks.

In order to further validate this work, a fine-grained human evaluation was conducted by annotators on the translation output in the EN$~\leftrightarrow~$GA directions and the findings are outlined in Linguistic Observations (cf. Section \ref{obs}).

As a multilingual tool, systems derived from adaptMLLM were also compared with the winning entries from the EN$~\leftrightarrow~$MR LoResMT2021 Shared Tasks. Fine-tuning 3.3B parameter NLLB models, using adaptMLLM demonstrated that our models for the EN$~\leftrightarrow~$MR language pair performed significantly better across all translation metrics when compared with the winning entries in the EN$~\leftrightarrow~$MR LoResMT2021 Shared Tasks.

The performance of our translation models developed for this study was compared with the output generated by Google Translate on both the EN$~\leftrightarrow~$GA and EN$~\leftrightarrow~$MR language pairs. In all language directions, the performance of the adaptMLLM models was better than that of Google Translate demonstrating a new SOTA in low-resource MT of the EN$~\leftrightarrow~$GA and EN$~\leftrightarrow~$MR language pairs. 

In terms of future work, there is much which can be performed to extend our research. There are several avenues which we plan on exploring further. Firstly, we would like to establish the effects of fine-tuning larger MLLMs, such as the 54B parameter NLLB network, on our existing datasets. It is anticipated this will most likely improve our results, and will also establish the trend in which increasingly larger MLLMs drive MT performance. The availability of the MTU and ADAPT GPU clusters, coupled with the deepspeed library, provides the platform upon which this can be achieved.

At this juncture, we have just scratched the surface of the MT performance enhancements which are possible through hyperparameter optimisation. Using a random search approach \cite{JMLR:v13:bergstra12a}, we will extend our search space by examining a greater number of hyperparameters and a larger range of associated values.   

Against this backdrop, it will be possible to apply adaptMLLM to new shared tasks and WMT competitions. This will also address another goal of our future work, which is to apply our approach to other low-resource language pairs.

Furthermore, integration of GPT-3, GPT-4, and BARD (cf. Section \ref{llm_background}) playgrounds into adaptMLLM, in addition to fine-tuning of these LLMs, will be explored in the future. 

Once the preserve of large research teams with very significant compute infrastructure, our approach has shown it is possible for much smaller research teams to fine-tune MLLMs on modest budgets. In doing so, we have succeeded in developing SOTA results for two low-resource language pairs. As an open-source initiative, we look forward to the community contributing to its advancement through the addition of fresh concepts and feature enhancements.

We have shown in the context of our low-resourced EN$~\leftrightarrow~$MR and EN$~\leftrightarrow~$GA pairs that fine-tuning a pre-trained MLLM such as NLLB is a more efficient and effective approach than training a bespoke Transformer model from scratch. 

In addition to improved performance, fine-tuning MLLM saves both time and computational resources. Consequently, given the right infrastructure, we recommend using such an approach when developing MT systems for low-resource pairs in the future.

\section{Limitations of the Study}

With additional resources, some elements of this research could be expanded upon. While there is a satisfactory level of agreement between annotators, the inclusion of a larger pool of annotators would be beneficial. Moreover, evaluating a more extensive selection of lines with a finer classification of the MQM taxonomy could yield deeper understanding of the MT outputs.

Whereas fine-tuning the baseline NLLB models highlighted a demonstrable improvement in translation quality using automatic metrics, a corresponding human evaluation of the baseline NLLB outputs was not conducted. As part of our future work, it is planned to conduct such an evaluation.

The focus of the study primarily centred on fine-tuning the NLLB base model, since it was the most likely candidate for success in producing high quality MT output for low-resource languages. Other LLMs, such as GPT-J, should also be investigated for fine-tuning~experiments.

With more hardware resources, and a larger research team, the impact of even larger models such as NLLB-54B would have been explored. It is planned to address these limitations in our future work (cf. Section \ref{concl}).

\vspace{6pt}

\authorcontributions{Writing—original draft, S.L.; Writing—review \& editing, H.A. and A.W. All authors have read and agreed to the published version of the manuscript.} 

\funding{This research is supported by Science Foundation Ireland through ADAPT Centre (Grant 13/RC/2106) (\url{https://www.adaptcentre.ie}, accessed on 28 November 2023) at Dublin City University. This research was also funded by the Staff Doctorate Scheme at the Munster Technological University.} 

\institutionalreview{In the “Related Work” section of this paper, we discuss academic papers published at conferences and in academic journals. We ensure that all data used in our analysis were obtained legally and ethically. With regard to licensing for our application, adaptMLLM, it is covered by the Creative Commons Attribution 4.0 International License. We recognise the importance of responsible and ethical conduct in AI research, and will continue to prioritise these values in our work.}

\dataavailability{The data presented in this study are openly available and can be found at \url{https://github.com/adaptNMT/adaptMLLM/} (accessed on 28 November 2023).}

\acknowledgments{ We also thank our anonymous reviewers for their comments, and our annotators Darragh Lankford and Muireann Ní Chorcora for their meticulous work in annotating the system outputs.}

\conflictsofinterest{the authors declare no conflict of interest. The funders had no role in the design of the study; in the collection, analyses, or interpretation of data; in the writing of the manuscript; or in the decision to publish the results.} 
 
\begin{adjustwidth}{-\extralength}{0cm}
\reftitle{References}



\PublishersNote{}
\end{adjustwidth}

\end{document}